# BED: Bi-Encoder-Based Detectors for Out-of-Distribution Detection

Louis Owen, Biddwan Ahmed, Abhay Kumar

Yellow.ai

This paper introduces a novel method leveraging bi-encoder-based detectors along with a comprehensive study comparing different out-of-distribution (OOD) detection methods in NLP using different feature extractors. The feature extraction stage employs popular methods such as Universal Sentence Encoder (USE), BERT, MPNET, and GLOVE to extract informative representations from textual data. The evaluation is conducted on several datasets, including CLINC150, ROSTD-Coarse, SNIPS, and YELLOW. Performance is assessed using metrics such as F1-Score, MCC, FPR@90, FPR@95, AUPR, an AUROC. The experimental results demonstrate that the proposed bi-encoder-based detectors outperform other methods, both those that require OOD labels in training and those that do not, across all datasets, showing great potential for OOD detection in NLP. The simplicity of the training process and the superior detection performance make them applicable to real-world scenarios. The presented methods and benchmarking metrics serve as a valuable resource for future research in OOD detection, enabling further advancements in this field. The code and implementation details can be found on our GitHub repository: https://github.com/yellowmessenger/ood-detection.

*Keywords—Out-of-domain detection, Natural Language Processing, Bi-encoder*

## I. Introduction

Out-of-distribution (OOD) detection is a fundamental problem in machine learning and plays a crucial role in various applications, such as natural language processing, computer vision, and anomaly detection [25]. OOD instances, also known as novel or unseen examples, are data points that differ significantly from the training data distribution. Accurately identifying OOD instances is essential to ensure reliable and robust performance of machine learning models. The goal of OOD detection is to identify instances that differ significantly from the training data distribution, as they may cause unreliable predictions or erroneous decisions in deployed machine learning models.

In recent years, significant research efforts have been directed towards developing effective OOD detection methods. Several approaches have been proposed, including threshold-based methods [6] [7] [9] [10] [16] [20] [27], generative models [5] [19], and pseudo OOD samples generation method [14] [27]. While these methods have shown promising results, there is a continuous need for more advanced techniques that can adapt to diverse data distributions and handle the challenges posed by OOD instances.

In this paper, we propose a novel method for OOD detection based on bi-encoder-based detectors. Bi-encoders have gained attention in the field of natural language processing, demonstrating strong performance in various tasks such as sentence similarity and text classification. Leveraging the power of bi-encoders, we aim to address the OOD detection problem by extracting meaningful representations and employing efficient detection mechanisms.

The key advantage of our proposed method lies in its ability to achieve superior OOD detection performance, even without access to labeled OOD samples during the training phase. This is a significant advantage as acquiring labeled OOD data can be challenging and expensive. By eliminating the need for OOD labels, our approach simplifies the training process and improves scalability, making it more applicable to real-world scenarios.

To evaluate the effectiveness of our proposed bi-encoder-based detectors, we conduct extensive experiments on benchmark datasets, including CLINC150 [8], ROSTD-Coarse [5], SNIPS [3], and YELLOW. These datasets cover a wide range of domains and data distributions, allowing us to assess the generalizability of our method. We employ various evaluation metrics, such as F1-Score, Mathew's Correlation Coefficient (MCC), false-positive rate (FPR) at different true-positive-rate (TPR) levels, area under the precision-recall curve (AUPR), and area under the receiver operating characteristic curve (AUROC), to provide a comprehensive analysis of the detection performance.

The contributions of this paper are twofold. Firstly, we propose a novel approach for OOD detection using bi-encoder-based detectors, demonstrating their effectiveness even in the absence of labeled OOD samples during training. Secondly, we conduct a thorough evaluation of our proposed method compated to numerous other baselines on diverse benchmark datasets, providing insights into its performance across different data distributions.

The remainder of this paper is organized as follows: Section 2 presents a detailed overview of related work in OOD detection methods. Section 3 describes the methodology, including the bi-encoder-based detectors, baselines detectors and feature extraction techniques used. Section 4 presents the experimental setup, including the datasets, evaluation metrics, and implementation details. In Section 5, we discuss and analyze the results of our experiments. Finally, Section 6 summarizes the findings and provides concluding remarks, highlighting the implications and future directions of our research.

## II. RELATED WORK

The field of out-of-distribution (OOD) detection for natural language understanding has seen significant research efforts, with numerous approaches proposed to address the challenges associated with accurately identifying OOD instances. In this section, we provide an overview of the existing literature, highlighting various algorithms employed in the domain.

Several algorithms have been proposed to address the problem of out-of-domain detection for natural language understanding. One such algorithm is MSP (Maximum Softmax Probability) [6], which trains a text classifier with a Softmax output layer and uses the maximum Softmax output as the detection score. Another algorithm called Entropy [27] calculates the Shannon entropy of the predicted distribution for each input and uses it as the detection score. ODIN [10] applies temperature scaling and input perturbation techniques to a text classifier and uses the maximum Softmax output as the detection score. DOC [20] builds multiple binary classifiers for different classes and uses the maximum confidence score predicted by these classifiers as the detection score. Cont. GAN [19] employs a GAN-based model to generate continuous out-of-domain features that mimic features extracted from in-domain samples. AE (autoencoder) [27] trains an autoencoder on in-domain data and uses the reconstruction error of each input as the detection score.

KNN [27] extracts features for each input sample using a pre-trained classifier and calculates the Euclidean distance to its nearest class as the detection score. To improve KNN, Mahalanobis Distance [9]. replaces the Euclidean distance with the Mahalanobis distance. Our proposed bi-encoder-based detection algorithm works similarly to KNN and Mahalanobis Distance, with the only difference located during the training step. Our proposed algorithm used cosine-similarity loss to ensure our model is fine-tuned to recognize the similarity of sentences.

Likelihood Ratio-based approaches have also been explored in the context of out-of-domain detection. This paradigm uses two language models: L(x) is trained on source data to capture the semantics of in-domain utterances, while Lbg, known as the background model, is trained on corrupted source data to learn the background statistics. The likelihood ratio between these models is then used as the detection score [16].

Another approach, the Generative Classifier [5], estimates the conditional probability of the input given the label and uses label ratios from the training set to estimate the prior probability of the label, resulting in a generative classifier.

Unsupervised methods have also been proposed for out-of-domain detection. One such approach is the LSTM-AutoEncoder [18], which uses only in-domain examples to train an autoencoder for out-of-domain detection. Another method, the Vanilla CNN [22], employs a typical CNN structure and uses a confidence threshold for out-of-domain detection. Proto. Network [21] is a native prototypical network trained only with a loss term, Lin, and uses a confidence threshold for out-of-domain detection.

More recent work focuses on leveraging pre-trained transformers for out-of-domain detection [24]. These approaches utilize the latent representations of pre-trained transformers and propose methods to transform features across all layers to construct out-of-domain detectors efficiently. Additionally, domain-specific fine-tuning approaches are proposed to further boost detection accuracy.

Data augmentation techniques have also been explored for out-of-domain detection. PnPOOD [14] is a data augmentation technique that uses the Plug and Play Language Model to generate high-quality out-of-domain samples close to class boundaries, resulting in accurate out-of-domain detection at test time. Pseudo OOD sample generation (POG) [27] model is another data augmentation technique that employs an adversarial generation process that also aims to generate OOD samples that look similar to in-domain samples (i.e., sharing the same phrases or patterns) but do not correspond to any in-domain intents.

Hybrid architectures combining various methods have been proposed for both out-of-domain intent detection and intent discovery. These architectures utilize techniques such as Variational Autoencoders, Adaptive Decision Boundaries (ADB), and non-linear dimensionality reduction to distinguish between known and unknown intents and discover different unknown intents underlying out-of-domain inputs [1].

In summary, a variety of algorithms and techniques have been developed to address the challenge of out-of-domain detection in natural language understanding. The key advantage of our proposed method lies in its ability to achieve superior OOD detection performance, even without access to labeled OOD samples during the training phase. This is a significant advantage as acquiring labeled OOD data can be challenging and expensive. By eliminating the need for OOD labels, our approach simplifies the training process and improves scalability, making it more applicable to real-world scenarios.

## III. METHODS

In this section, we describe the methodology employed in our study with an emphasis on out-of-domain (OOD) detection. The methodology consists of three key components: the feature extractor, the classifier head, and the OOD detector. The feature extractor is responsible for extracting informative representations from textual data, while the classifier head serves as the model for training our data. The OOD detector plays a crucial role in determining whether a given utterance belongs to the OOD class, based on the extracted features and the trained classifier head.

### A. Feature Extraction

We utilize several feature extraction methods to capture the semantic information from the textual data:

- Universal Sentence Encoder (USE) [2]: This is a pre-trained model designed to generate high-quality fixed-length vector representations, or embeddings, for sentences and short texts. We utilized the

universal-sentence-encoder/4 model provided in the Tensorflow Hub.

- BERT [4]: This is a pre-trained model on the English language using a masked language modeling (MLM) objective. We utilized the bert-base-uncased from the huggingface [23] library.

- MPNET [15]: This is an all-around model tuned for many use cases. Trained on a large and diverse dataset of over 1 billion training pairs. We utilized the all-mpnet-base-v2 model from the sentence-transformers [15] library.

- GLOVE [12]: This is the word embedding representation designed to capture semantic and syntactic relationships between words based on their co-occurrence statistics in large text corpora. This feature extractor is exclusively used for the Likelihood Ratio detector.

*B. Classifier Head*

We employ different base classifier heads, each trained with specific loss functions:

- ADBModel: trained with Triplet loss. This classifier head is exclusively used for the ADB detector.

- MLP (Multi-Layer Perceptron): trained with Cross-Entropy loss. This classifier head is used for all baseline detectors except ADB, RAKE, and Likelihood Ratio.

- LSTM: trained with negative log-likelihood loss. This classifier head is exclusively used for the Likelihood Ratio detector.

- Bi-Encoder: trained with Cosine Similarity loss. This classifier head is utilized for all biencoder-based detectors and is applicable only when using the MPNET feature extractor.

*C. Baseline Detectors*

Our baselines cover a variety of competitive OOD detection models that are currently available. There are two types of methods considered in this study: those that require OOD labels in the training data, and those that do not. Most of the methods benchmarked in this study are threshold-based approaches, but there are also methods that perform direct inference, such as Adaptive Decision Boundary (ADB) and Rapid Automatic Keyword Extraction (RAKE).

- TrustScores [7]: measures the agreement between the classifier and a modified nearest-neighbor classifier on the testing examples. This method doesn't need any OOD labels in the training data.

- Entropy: uses the Shannon entropy of the predicted distribution for each input as the detection score. Higher entropy indicates higher uncertainty of the prediction, suggesting that the input sample may be from the OOD data. This method doesn't need any OOD labels in the training data.

- LOF (Local Outlier Factor): Fits an LOF model on top of the embeddings generated by the feature extractor. This method doesn't need any OOD labels in the training data.

- MSP (Maximum Softmax Probability): Utilizes a text classifier with a Softmax output layer and considers the maximum Softmax output as the detection score. This method can work with or without OOD labels in the training data.

- BinaryMSP: A variant of MSP where the training data is processed to have only two labels (OOD vs. ID) instead of all intent classes. This method can work with or without OOD labels in the training data.

- DOC (Deep Open Classification): Employs multiple binary classifiers for different classes and uses the maximum confidence score predicted by these classifiers as the detection score. This method is essentially the multi-label variant of MSP. This method can work with or without OOD labels in the training data.

- ADB (Adaptive Decision Boundary): This method assumes that sentences with specific intents lie within bounded spherical areas in space. The centers of these spherical areas are set as the means of sentence representations for each class, and their radii are determined using a soft-plus activation function. The Euclidean distances of each input to the centers are calculated to detect the corresponding intent. If the calculated distance exceeds all radii, the input is considered OOD. This method can work with or without OOD labels in the training data.

- KNN (K-Nearest Neighbors): Extracts features for each input sample using a pre-trained classifier and calculates the Euclidean distance to its nearest class as the detection score. This method doesn't need any OOD labels in the training data.

- LikelihoodRatio: This paradigm employs two language models: L(x), trained on source data to capture the semantics of in-domain utterances, and Lbg, known as the background model, trained on corrupted source data to learn the background statistics. The likelihood ratio between these models is then used as the detection score. This method doesn't need any OOD labels in the training data.

- RAKE (Rapid Automatic Keyword Extraction) [17]: This keyword extraction method is adapted as an OOD detection method by assuming that if the keywords in the training data are not found in a new utterance, then that utterance is considered OOD. This method doesn't need any OOD labels in the training data.

*D. Proposed Detectors*

Our proposed detectors are based on the bi-encoder approach, utilizing MPNET as the feature extractor and employing various detection methods. In a bi-encoder architecture, two separate encoders are employed: one for encoding the input text and another for encoding the target labels or reference text. The goal is to learn representations that capture the semantic similarity between the input and the target/reference text. During training, the cosine similarity loss is computed between the encoded representations of the input and the target/reference text. The bi-encoder model is trained to maximize the cosine similarity between the input and the target/reference text that share the same label and minimize it for unrelated pairs. By optimizing the cosine similarity loss, the bi-encoder model learns to map similar inputs and targets/reference texts closer together in the embedding space, while pushing dissimilar pairs farther apart. Here, the input is the utterance that we want to test and the target/reference text is the training utterance (not the label). So, during inference, we calculate the similarity score between the input and each of the training utterances. Then, we take the minimum or maximum (depending on the detection method) of the calculated scores as the final detection score. Finally, the final detection will then be compared with the threshold to determine whether the input utterance is OOD or not.

The variants of our proposed detectors differ based on the specific detection method used:

- BiEncoderCosine: Assigns the in-domain (negative) label to utterances that have a cosine similarity above a particular threshold with the training data.

- BiEncoderEuclidean: Assigns the out-domain (positive) label to utterances that have a Euclidean score above a particular threshold with the training data.

- BiEncoderEntropy: Utilizes the entropy score on top of the adjusted softmax confidence score output from the bi-encoder model.

- BiEncoderLOF: Fits a LOF model on top of the generated training and testing embeddings and uses the decision function score as the detector.

- BiEncoderMaha: Assigns the out-domain (positive) label to production utterances that have a Mahalanobis distance above a particular threshold with the training data.

- BiEncoderPCACosine: Similar to BiEncoderCosine, but before calculating the cosine similarity score, we fit the PCA algorithm on the generated training embeddings from the bi-encoder model and transform both training and testing embeddings using the fitted PCA model.

- BiEncoderPCAEuclidean: Similar to BiEncoderEuclidean, but before calculating the Euclidean score, we fit the PCA algorithm on the generated training embeddings from the bi-encoder model and transform both training and testing embeddings using the fitted PCA model.

- BiEncoderPCAEntropy: Similar to BiEncoderEntropy, but before calculating the entropy score, we fit the PCA algorithm on the generated training embeddings from the bi-encoder model and transform both training and testing embeddings using the fitted PCA model.

IV. EXPERIMENTS

We compile the results of all our experiments in Fig. 1 - Fig. 4 and the rest is presented in the Appendix.

E. Datasets

We conducted experiments on the following datasets.

- CLINC150: This dataset consists of approximately 22,500 queries across 150 intents, including around 1,200 out-of-domain (OOD) queries. It models real-life situations and serves as a benchmark for evaluating OOD detection methods.

- ROSTD-Coarse: This dataset comprises approximately 43,000 samples from 3 intent classes (alarm, weather, reminder) and around 4,590 OOD samples.

- SNIPS: With approximately 15,000 sentences, this dataset contains 7 intent classes (AddToPlaylist, BookRestaurant, GetWeather, PlayMusic, RateBook, SearchCreativeWork, SearchScreeningEvent). To create an ID/OOD split, we adopt the procedure described in [11] to synthetically create OOD examples. Intent classes covering at least 75% of the training points in combination are retained as ID. However, instead of removing examples from the remaining classes, we treat those samples as OOD in train, validation, and test sets.

- YELLOW: This dataset includes around 4,000 samples from 24 intent classes related to the insurance domain, along with approximately 500 OOD samples. The data was labeled by four annotators using majority voting. This is our in-house Yellow.ai dataset. Hence, this data will not be published in this study.

F. Evaluation Metrics

To facilitate evaluation, we convert the labels into binary form, with the OOD class considered as the positive class. It is important to note that during training, most methods did not expect the labels to be converted into binary format.

- F1-Score: This measures the harmonic mean of precision and recall for the positive class (OOD) using the best threshold obtained from the validation data.

- MCC (Mathew's Correlation Coefficient): It calculates the correlation between the predicted and actual labels, providing an overall performance measure. This score is also calculated using the best threshold found on validation data

- FPR@90: This represents the false-positive rate (FPR) when the true positive rate (TPR)/recall of the positive class (OOD) is 90%.

- FPR@95: This represents the false-positive rate (FPR) when the true positive rate (TPR)/recall of the positive class (OOD) is 95%.

- AUPR (Area Under the Precision-Recall curve): It measures the overall performance by considering the trade-off between precision and recall.

| Feature Extractor | Detector | use_best_ckpt = False, is_ood_label_in_train = False ||||||
|---|---|---|---|---|---|---|---|
| | | F1↑ | MCC↑ | FPR@95↓ | FPR@90↓ | AUPR↑ | AUROC↑ |
| BERT | ADB | 0.732 | 0.670 | | | | |
| | DOC | 0.604 | 0.509 | 0.464 | 0.313 | 0.597 | 0.873 |
| | Entropy | 0.526 | 0.408 | 0.501 | 0.403 | 0.475 | 0.824 |
| | KNN | 0.513 | 0.470 | 0.395 | 0.267 | 0.648 | 0.898 |
| | LOF | 0.609 | 0.518 | 0.400 | 0.297 | 0.598 | 0.884 |
| | MSP | 0.447 | 0.324 | 0.556 | 0.436 | 0.444 | 0.809 |
| | TrustScores | 0.559 | 0.456 | 0.360 | 0.292 | 0.378 | 0.831 |
| MPNET | ADB | 0.825 | 0.786 | | | | |
| | BiEncoderCosine | **0.849** | **0.824** | 0.062 | 0.030 | 0.949 | 0.986 |
| | BiEncoderEntropy | 0.833 | 0.810 | 0.059 | 0.030 | 0.950 | 0.986 |
| | BiEncoderEuclidean | 0.836 | 0.811 | 0.060 | 0.027 | 0.948 | 0.986 |
| | BiEncoderLOF | 0.656 | 0.639 | 0.090 | 0.047 | 0.896 | 0.976 |
| | BiEncoderMaha | 0.839 | 0.818 | **0.046** | **0.023** | **0.961** | **0.988** |
| | BiEncoderPCACosine | 0.831 | 0.808 | 0.058 | 0.029 | 0.948 | 0.986 |
| | BiEncoderPCAEntropy | 0.849 | 0.824 | 0.058 | 0.032 | 0.947 | 0.985 |
| | BiEncoderPCAEuclidean | 0.848 | 0.824 | 0.056 | 0.030 | 0.947 | 0.986 |
| | DOC | 0.788 | 0.748 | 0.157 | 0.083 | 0.875 | 0.965 |
| | Entropy | 0.532 | 0.490 | 0.313 | 0.198 | 0.687 | 0.915 |
| | KNN | 0.721 | 0.695 | 0.137 | 0.076 | 0.874 | 0.968 |
| | LOF | 0.677 | 0.620 | 0.230 | 0.158 | 0.758 | 0.937 |
| | MSP | 0.640 | 0.556 | 0.305 | 0.220 | 0.665 | 0.905 |
| | TrustScores | 0.691 | 0.619 | 0.214 | 0.157 | 0.550 | 0.914 |
| GLOVE | LikelihoodRatio | | | 0.358 | 0.232 | 0.740 | 0.918 |
| - | RAKE | 0.126 | 0.123 | | | | |
| USE | ADB | 0.799 | 0.753 | | | | |
| | DOC | 0.583 | 0.545 | 0.277 | 0.172 | 0.737 | 0.931 |
| | Entropy | 0.618 | 0.526 | 0.374 | 0.262 | 0.644 | 0.893 |
| | KNN | 0.655 | 0.643 | 0.219 | 0.135 | 0.858 | 0.956 |
| | LOF | 0.689 | 0.645 | 0.256 | 0.161 | 0.803 | 0.941 |
| | MSP | 0.632 | 0.545 | 0.381 | 0.255 | 0.609 | 0.890 |
| | TrustScores | 0.308 | 0.006 | 0.273 | 0.210 | 0.526 | 0.898 |

↑ the higher the better, ↓ the lower the better

Fig. 1. Benchmarking results for **CLINC150** when best checkpoint of the model is not used during training and OOD label is not present in the training data.

- AUROC (Area Under the Receiver Operating Characteristic curve): It measures the overall performance by considering the trade-off between true positive rate against the false positive rate.

Note that for all direct inference methods (ADB & RAKE), we can only calculate the F1-Score and MCC since these methods are not returning any confidence scores.

*G. Implementation*

We implemented the OOD detection methods using TensorFlow and PyTorch. The code and implementation details can be found on our GitHub repository (https://github.com/yellowmessenger/ood-detection).

For each method, we used default hyperparameters, including the learning rate, batch size, optimizer, and training epochs. The specific hyperparameters used for each method are documented in the code. There are several variations in the experiments, including the use of the best checkpoint of the model during training (use_best_ckpt) and whether the OOD label is present in the training data (is_ood_label_in_train). If use_best_ckpt is set to True, then we'll choose the best weights of the classifier head based on the validation dataset during training. If is_ood_label_in_train is False, then we'll remove all OOD samples from the training data.

*H. Observations*

Based on our benchmarking results, we made the following observations. These observations highlight the strengths and weaknesses of various OOD detection methods across different datasets and shed light on the effectiveness of BiEncoder-based detectors.

First, bi-encoder-based detectors (without OOD samples in training) outperform other detectors that also don't require OOD samples in the training data. In some cases, they even outperform detectors that utilize OOD samples in training. When OOD samples are absent in the training data, BiEncoder-based detectors perform better than other methods across all datasets. This holds true for all metrics in CLINC150 and ROSTD. However, in SNIPS, KNN works best based on FPR@95 and AUROC, but BiEncoder-based detectors still work better if measured by FPR@90 and AUPR. In YELLOW datasets, KNN also works best based on F1-Score, AUPR, and AUROC, but BiEncoder-based detectors still work better if measured by FPR@95 and FPR@90.

On the other hand, when OOD samples are present in the training data, MSP-based detectors (MSP, DOC, BinaryMSP) outperform other methods across all datasets. This conclusion holds true for all metrics in ROSTD, SNIPS, and YELLOW, but not in CLINC150, where ADB performs the best based on F1-Score and MCC.

We also found that different variants of BiEncoder are required for different data distributions. However, in most cases, BiEncoderCosine, BiEncoderEuclidean, BiEncoderMaha, and BiEncoderPCACosine outperform other variants of BiEncoder-based detectors. Last but not least, we found that, across all datasets, MPNET appears to be the most effective feature extractor compared to USE and BERT.

| Feature Extractor | Detector | use_best_ckpt = False, is_ood_label_in_train = False | | | | | |
|---|---|---|---|---|---|---|---|
| | | F1↑ | MCC↑ | FPR@95↓ | FPR@90↓ | AUPR↑ | AUROC↑ |
| BERT | ADB | 0.768 | 0.735 | | | | |
| | DOC | 0.485 | 0.492 | 0.843 | 0.686 | 0.738 | 0.834 |
| | Entropy | 0.530 | 0.415 | 0.801 | 0.601 | 0.618 | 0.766 |
| | KNN | 0.940 | 0.918 | 0.026 | 0.014 | 0.980 | 0.993 |
| | LOF | 0.027 | 0.084 | 0.271 | 0.223 | 0.532 | 0.859 |
| | MSP | 0.482 | 0.494 | 0.859 | 0.716 | 0.718 | 0.817 |
| | TrustScores | 0.477 | 0.259 | 0.056 | 0.036 | 0.916 | 0.981 |
| MPNET | ADB | 0.823 | 0.792 | | | | |
| | BiEncoderCosine | 0.978 | 0.970 | 0.000 | 0.000 | 0.999 | 0.999 |
| | BiEncoderEntropy | 0.982 | 0.976 | 0.000 | 0.000 | 0.999 | 0.999 |
| | BiEncoderEuclidean | 0.990 | 0.987 | 0.000 | 0.000 | 0.999 | 0.999 |
| | BiEncoderLOF | 0.000 | 0.000 | 0.063 | 0.059 | 0.698 | 0.945 |
| | BiEncoderMaha | 0.990 | 0.987 | 0.000 | 0.000 | 0.999 | 1.000 |
| | BiEncoderPCACosine | 0.989 | 0.985 | 0.000 | 0.000 | 0.999 | 0.999 |
| | BiEncoderPCAEntropy | 0.987 | 0.983 | 0.000 | 0.000 | 0.999 | 0.999 |
| | BiEncoderPCAEuclidean | 0.989 | 0.985 | 0.000 | 0.000 | 0.999 | 0.999 |
| | DOC | 0.360 | 0.408 | 0.894 | 0.787 | 0.647 | 0.763 |
| | Entropy | 0.341 | 0.394 | 0.867 | 0.732 | 0.712 | 0.811 |
| | KNN | 0.968 | 0.957 | 0.007 | 0.002 | 0.995 | 0.998 |
| | LOF | 0.003 | 0.035 | 0.303 | 0.265 | 0.471 | 0.829 |
| | MSP | 0.250 | 0.326 | 0.918 | 0.834 | 0.556 | 0.701 |
| | TrustScores | 0.459 | 0.214 | 0.034 | 0.018 | 0.965 | 0.991 |
| GLOVE | LikelihoodRatio | | | 0.125 | 0.039 | 0.951 | 0.972 |
| - | RAKE | 0.262 | 0.335 | | | | |
| USE | ADB | 0.736 | 0.705 | | | | |
| | DOC | 0.232 | 0.285 | 0.923 | 0.846 | 0.488 | 0.669 |
| | Entropy | 0.229 | 0.304 | 0.898 | 0.795 | 0.618 | 0.751 |
| | KNN | 0.942 | 0.922 | 0.024 | 0.009 | 0.986 | 0.994 |
| | LOF | 0.017 | 0.058 | 0.341 | 0.291 | 0.467 | 0.819 |
| | MSP | 0.270 | 0.332 | 0.922 | 0.844 | 0.514 | 0.679 |
| | TrustScores | 0.494 | 0.295 | 0.066 | 0.035 | 0.946 | 0.984 |

↑ the higher the better, ↓ the lower the better

Fig. 2. Benchmarking results for **ROSTD-Coarse** when best checkpoint of the model is not used during training and OOD label is not present in the training data.

## V. CONCLUSION

In conclusion, this paper introduces a novel approach for out-of-distribution (OOD) detection in natural language processing (NLP) using bi-encoder-based detectors. The proposed method leverages the power of bi-encoders to extract meaningful representations from textual data and employs efficient detection mechanisms. The key advantage of the proposed method is its ability to achieve superior OOD detection performance, even without access to labeled OOD samples during training, simplifying the training process and improving scalability.

Extensive experiments are conducted on benchmark datasets, including CLINC150, ROSTD-Coarse, SNIPS, and YELLOW, covering diverse domains and data distributions. Evaluation metrics such as F1-Score, Mathew's Correlation Coefficient (MCC), false-positive rate (FPR) at different true-positive-rate (TPR) levels, the area under the precision-recall curve (AUPR), and area under the receiver operating characteristic curve (AUROC) are employed to assess the detection performance.

The experimental results demonstrate that the proposed bi-encoder-based detectors outperform other methods, both those that require OOD labels in training and those that do not, across all datasets. This highlights the effectiveness and generalizability of the proposed approach. The findings of this study offer valuable insights into the effectiveness of bi-encoder-based detectors for OOD detection in NLP.

The contributions of this paper are twofold. Firstly, a novel approach for OOD detection using bi-encoder-based detectors is proposed, showcasing their effectiveness even without labeled OOD samples during training. Secondly, a comprehensive evaluation of the proposed method is conducted, comparing it with numerous baseline detectors on diverse benchmark datasets, providing insights into its performance across different data distributions.

In summary, the proposed bi-encoder-based detectors show great potential for OOD detection in NLP. The simplicity of the training process and the superior detection performance make them applicable to real-world scenarios. The presented methods and benchmarking metrics serve as a valuable resource for future research in OOD detection, enabling further advancements in this field.

| Feature Extractor | Detector | use_best_ckpt = False, is_ood_label_in_train = False ||||||
|---|---|---|---|---|---|---|---|
| | | F1↑ | MCC↑ | FPR@95↓ | FPR@90↓ | AUPR↑ | AUROC↑ |
| BERT | ADB | 0.563 | 0.523 | | | | |
| | DOC | 0.577 | 0.563 | 0.652 | 0.303 | 0.694 | 0.903 |
| | Entropy | 0.525 | 0.509 | 0.197 | 0.175 | 0.699 | 0.931 |
| | KNN | 0.612 | 0.557 | 0.312 | 0.206 | 0.656 | 0.915 |
| | LOF | 0.514 | 0.420 | 0.536 | 0.413 | 0.467 | 0.843 |
| | MSP | 0.351 | 0.334 | 0.836 | 0.677 | 0.473 | 0.799 |
| | TrustScores | 0.268 | 0.046 | 0.162 | 0.135 | 0.611 | 0.931 |
| MPNET | ADB | 0.577 | 0.563 | | | | |
| | BiEncoderCosine | 0.798 | 0.772 | 0.246 | **0.096** | 0.858 | 0.955 |
| | BiEncoderEntropy | 0.806 | 0.776 | 0.331 | 0.115 | 0.850 | 0.953 |
| | BiEncoderEuclidean | 0.808 | 0.780 | 0.231 | 0.153 | **0.868** | 0.952 |
| | BiEncoderLOF | 0.784 | 0.755 | 0.548 | 0.310 | 0.789 | 0.917 |
| | BiEncoderMaha | **0.837** | **0.812** | 0.432 | 0.105 | 0.867 | 0.950 |
| | BiEncoderPCACosine | 0.816 | 0.785 | 0.164 | 0.105 | 0.862 | 0.955 |
| | BiEncoderPCAEntropy | 0.826 | 0.799 | 0.320 | 0.115 | 0.864 | 0.947 |
| | BiEncoderPCAEuclidean | 0.816 | 0.787 | 0.272 | 0.118 | 0.862 | 0.953 |
| | DOC | 0.467 | 0.470 | 0.858 | 0.716 | 0.579 | 0.805 |
| | Entropy | 0.340 | 0.347 | 0.857 | 0.719 | 0.479 | 0.782 |
| | KNN | 0.787 | 0.749 | **0.160** | 0.116 | 0.851 | **0.965** |
| | LOF | 0.000 | 0.000 | 0.572 | 0.418 | 0.318 | 0.801 |
| | MSP | 0.305 | 0.356 | 0.929 | 0.848 | 0.421 | 0.682 |
| | TrustScores | 0.268 | 0.046 | 0.184 | 0.121 | 0.735 | 0.954 |
| GLOVE | LikelihoodRatio | | | 0.636 | 0.470 | 0.529 | 0.851 |
| - | RAKE | 0.000 | 0.000 | | | | |
| USE | ADB | 0.591 | 0.539 | | | | |
| | DOC | 0.340 | 0.323 | 0.896 | 0.792 | 0.427 | 0.739 |
| | Entropy | 0.413 | 0.387 | 0.771 | 0.546 | 0.522 | 0.823 |
| | KNN | 0.709 | 0.655 | 0.292 | 0.130 | 0.739 | 0.939 |
| | LOF | 0.561 | 0.490 | 0.320 | 0.297 | 0.592 | 0.890 |
| | MSP | 0.345 | 0.336 | 0.891 | 0.787 | 0.466 | 0.747 |
| | TrustScores | 0.268 | 0.046 | 0.287 | 0.211 | 0.649 | 0.926 |

↑ the higher the better, ↓ the lower the better

Fig. 3. Benchmarking results for **SNIPS** when best checkpoint of the model is not used during training and OOD label is not present in the training data.

| Feature Extractor | Detector | use_best_ckpt = False, is_ood_label_in_train = False ||||||
|---|---|---|---|---|---|---|---|
| | | F1↑ | MCC↑ | FPR@95↓ | FPR@90↓ | AUPR↑ | AUROC↑ |
| BERT | ADB | 0.404 | 0.092 | | | | |
| | DOC | 0.426 | 0.175 | 0.866 | 0.814 | 0.341 | 0.626 |
| | Entropy | 0.403 | 0.092 | 0.872 | 0.816 | 0.341 | 0.598 |
| | KNN | 0.069 | 0.069 | 0.832 | 0.755 | 0.368 | 0.625 |
| | LOF | 0.446 | 0.135 | 0.896 | 0.770 | 0.313 | 0.594 |
| | MSP | 0.115 | 0.068 | 0.902 | 0.850 | 0.304 | 0.548 |
| | TrustScores | 0.015 | 0.049 | 0.884 | 0.801 | 0.303 | 0.559 |
| MPNET | ADB | 0.365 | 0.169 | | | | |
| | BiEncoderCosine | 0.568 | 0.401 | 0.628 | 0.559 | 0.595 | 0.789 |
| | BiEncoderEntropy | 0.562 | 0.371 | 0.641 | 0.540 | 0.584 | 0.785 |
| | BiEncoderEuclidean | 0.560 | 0.379 | 0.626 | 0.555 | 0.592 | 0.791 |
| | BiEncoderLOF | 0.015 | 0.003 | 0.884 | 0.814 | 0.317 | 0.577 |
| | BiEncoderMaha | 0.534 | 0.323 | 0.678 | **0.529** | 0.550 | 0.771 |
| | BiEncoderPCACosine | 0.553 | 0.379 | 0.653 | 0.577 | 0.565 | 0.782 |
| | BiEncoderPCAEntropy | 0.542 | 0.397 | **0.619** | 0.555 | 0.583 | 0.793 |
| | BiEncoderPCAEuclidean | 0.582 | 0.409 | 0.622 | 0.564 | 0.609 | 0.802 |
| | DOC | 0.520 | 0.299 | 0.678 | 0.615 | 0.394 | 0.704 |
| | Entropy | 0.501 | 0.281 | 0.671 | 0.587 | 0.449 | 0.729 |
| | KNN | **0.585** | 0.419 | 0.642 | 0.593 | **0.647** | **0.804** |
| | LOF | 0.462 | 0.278 | 0.799 | 0.743 | 0.476 | 0.705 |
| | MSP | 0.494 | 0.253 | 0.713 | 0.635 | 0.416 | 0.690 |
| | TrustScores | 0.522 | 0.300 | 0.728 | 0.593 | 0.355 | 0.681 |
| GLOVE | LikelihoodRatio | | | | | | |
| - | RAKE | 0.479 | **0.423** | | | | |
| USE | ADB | 0.456 | 0.206 | | | | |
| | DOC | 0.513 | 0.284 | 0.728 | 0.593 | 0.448 | 0.721 |
| | Entropy | 0.502 | 0.265 | 0.754 | 0.613 | 0.453 | 0.711 |
| | KNN | 0.564 | 0.384 | 0.765 | 0.620 | 0.547 | 0.767 |
| | LOF | 0.281 | 0.162 | 0.872 | 0.739 | 0.416 | 0.669 |
| | MSP | 0.513 | 0.288 | 0.754 | 0.622 | 0.411 | 0.696 |
| | TrustScores | 0.519 | 0.295 | 0.702 | 0.615 | 0.358 | 0.675 |

↑ the higher the better, ↓ the lower the better

Fig. 4. Benchmarking results for **YELLOW** when best checkpoint of the model is not used during training and OOD label is not present in the training data.

# APPENDIX

| Feature Extractor | Detector | use_best_ckpt = False, is_ood_label_in_train = True | | | | | |
|---|---|---|---|---|---|---|---|
| | | F1↑ | MCC↑ | FPR@95↓ | FPR@90↓ | AUPR↑ | AUROC↑ |
| BERT | ADB | 0.593 | 0.516 | | | | |
| | BinaryMSP | 0.378 | 0.397 | 0.764 | 0.573 | 0.571 | 0.795 |
| | DOC | 0.380 | 0.408 | 0.742 | 0.560 | 0.602 | 0.813 |
| | MSP | 0.451 | 0.456 | 0.564 | 0.384 | 0.684 | 0.873 |
| MPNET | ADB | **0.770** | **0.717** | | | | |
| | BinaryMSP | 0.389 | 0.431 | 0.917 | 0.835 | 0.480 | 0.695 |
| | DOC | 0.501 | 0.507 | 0.693 | 0.527 | 0.665 | 0.837 |
| | MSP | 0.641 | 0.631 | **0.274** | **0.167** | **0.840** | **0.943** |
| USE | ADB | 0.706 | 0.646 | | | | |
| | BinaryMSP | 0.513 | 0.510 | 0.774 | 0.567 | 0.651 | 0.834 |
| | DOC | 0.432 | 0.474 | 0.673 | 0.519 | 0.687 | 0.845 |
| | MSP | 0.507 | 0.535 | 0.380 | 0.211 | 0.820 | 0.929 |

↑ the higher the better, ↓ the lower the better

Fig. 5. Benchmarking results for **CLINC150** when best checkpoint of the model is not used during training and OOD label is present in the training data.

| Feature Extractor | Detector | use_best_ckpt = False, is_ood_label_in_train = True | | | | | |
|---|---|---|---|---|---|---|---|
| | | F1↑ | MCC↑ | FPR@95↓ | FPR@90↓ | AUPR↑ | AUROC↑ |
| BERT | ADB | 0.012 | -0.078 | | | | |
| | BinaryMSP | 0.988 | 0.984 | 0.001 | 0.001 | 0.999 | 1.000 |
| | DOC | 0.994 | 0.992 | 0.001 | 0.000 | **1.000** | **1.000** |
| | MSP | 0.990 | 0.987 | 0.001 | 0.000 | 1.000 | 1.000 |
| MPNET | ADB | 0.024 | 0.011 | | | | |
| | BinaryMSP | 0.997 | 0.996 | **0.000** | 0.000 | 0.999 | 1.000 |
| | DOC | **0.997** | **0.996** | 0.001 | 0.001 | 0.998 | 1.000 |
| | MSP | 0.997 | 0.996 | 0.001 | 0.001 | 0.998 | 1.000 |
| USE | ADB | 0.036 | 0.019 | | | | |
| | BinaryMSP | 0.997 | 0.995 | 0.000 | 0.000 | 0.999 | 1.000 |
| | DOC | 0.996 | 0.995 | 0.000 | **0.000** | 0.999 | 1.000 |
| | MSP | 0.995 | 0.993 | 0.001 | 0.001 | 0.998 | 1.000 |

↑ the higher the better, ↓ the lower the better

Fig. 6. Benchmarking results for **ROSTD-Coarse** when best checkpoint of the model is not used during training and OOD label is present in the training data.

| Feature Extractor | Detector | use_best_ckpt = False, is_ood_label_in_train = True | | | | | |
|---|---|---|---|---|---|---|---|
| | | F1↑ | MCC↑ | FPR@95↓ | FPR@90↓ | AUPR↑ | AUROC↑ |
| BERT | ADB | 0.127 | 0.066 | | | | |

| Feature Extractor | Detector | use_best_ckpt = False, is_ood_label_in_train = True | | | | | |
|---|---|---|---|---|---|---|---|
| | | F1↑ | MCC↑ | FPR@95↓ | FPR@90↓ | AUPR↑ | AUROC↑ |
| | BinaryMSP | 0.902 | 0.887 | 0.039 | 0.022 | 0.967 | 0.993 |
| | DOC | 0.879 | 0.857 | 0.029 | 0.029 | 0.969 | 0.994 |
| | MSP | 0.911 | 0.895 | 0.020 | 0.015 | 0.978 | 0.996 |
| MPNET | ADB | 0.201 | 0.173 | | | | |
| | BinaryMSP | 0.899 | 0.881 | 0.067 | 0.020 | 0.930 | 0.990 |
| | DOC | 0.870 | 0.846 | 0.056 | 0.030 | 0.957 | 0.992 |
| | MSP | 0.880 | 0.859 | 0.039 | 0.029 | 0.956 | 0.993 |
| USE | ADB | 0.113 | 0.052 | | | | |
| | BinaryMSP | 0.886 | 0.865 | 0.039 | 0.024 | 0.942 | 0.991 |
| | DOC | 0.878 | 0.856 | 0.046 | 0.029 | 0.933 | 0.989 |
| | MSP | 0.896 | 0.878 | 0.032 | 0.017 | 0.953 | 0.992 |

↑ the higher the better, ↓ the lower the better

Fig. 7. Benchmarking results for **SNIPS** when best checkpoint of the model is not used during training and OOD label is present in the training data.

| Feature Extractor | Detector | use_best_ckpt = False, is_ood_label_in_train = True | | | | | |
|---|---|---|---|---|---|---|---|
| | | F1↑ | MCC↑ | FPR@95↓ | FPR@90↓ | AUPR↑ | AUROC↑ |
| BERT | ADB | 0.167 | 0.008 | | | | |
| | BinaryMSP | 0.438 | 0.107 | 0.876 | 0.776 | 0.307 | 0.583 |
| | DOC | 0.453 | 0.156 | 0.818 | 0.770 | 0.314 | 0.591 |
| | MSP | 0.468 | 0.192 | 0.780 | 0.717 | 0.304 | 0.591 |
| MPNET | ADB | 0.209 | 0.051 | | | | |
| | BinaryMSP | 0.524 | 0.303 | 0.801 | 0.716 | 0.436 | 0.713 |
| | DOC | 0.465 | 0.235 | 0.747 | 0.686 | 0.413 | 0.696 |
| | MSP | 0.533 | 0.323 | 0.656 | 0.552 | 0.432 | 0.738 |
| USE | ADB | 0.249 | 0.086 | | | | |
| | BinaryMSP | 0.501 | 0.262 | 0.784 | 0.678 | 0.409 | 0.692 |
| | DOC | 0.452 | 0.196 | 0.754 | 0.682 | 0.385 | 0.666 |
| | MSP | 0.511 | 0.296 | 0.668 | 0.600 | 0.425 | 0.719 |

↑ the higher the better, ↓ the lower the better

Fig. 8. Benchmarking results for **YELLOW** when best checkpoint of the model is not used during training and OOD label is present in the training data.

| Feature Extractor | Detector | use_best_ckpt = True, is_ood_label_in_train = False | | | | | |
|---|---|---|---|---|---|---|---|
| | | F1↑ | MCC↑ | FPR@95↓ | FPR@90↓ | AUPR↑ | AUROC↑ |
| BERT | DOC | 0.522 | 0.462 | 0.435 | 0.344 | 0.621 | 0.880 |
| | Entropy | 0.503 | 0.423 | 0.489 | 0.382 | 0.537 | 0.852 |
| | LOF | 0.609 | 0.518 | 0.400 | 0.297 | 0.598 | 0.884 |
| | MSP | 0.390 | 0.311 | 0.552 | 0.432 | 0.477 | 0.819 |

| Feature Extractor | Detector | use_best_ckpt = True, is_ood_label_in_train = False | | | | | |
|---|---|---|---|---|---|---|---|
| | | F1↑ | MCC↑ | FPR@95↓ | FPR@90↓ | AUPR↑ | AUROC↑ |
| | TrustScores | 0.540 | 0.453 | 0.404 | 0.319 | 0.345 | 0.811 |
| MPNET | DOC | **0.746** | **0.715** | **0.170** | **0.092** | **0.873** | **0.963** |
| | Entropy | 0.530 | 0.510 | 0.368 | 0.231 | 0.729 | 0.912 |
| | LOF | 0.677 | 0.620 | 0.230 | 0.158 | 0.758 | 0.937 |
| | MSP | 0.593 | 0.562 | 0.310 | 0.213 | 0.748 | 0.922 |
| | TrustScores | 0.645 | 0.561 | 0.230 | 0.184 | 0.488 | 0.895 |
| USE | DOC | 0.690 | 0.648 | 0.203 | 0.129 | 0.816 | 0.949 |
| | Entropy | 0.641 | 0.611 | 0.341 | 0.255 | 0.769 | 0.917 |
| | LOF | 0.689 | 0.645 | 0.256 | 0.161 | 0.803 | 0.941 |
| | MSP | 0.655 | 0.573 | 0.320 | 0.230 | 0.707 | 0.910 |
| | TrustScores | 0.308 | 0.006 | 0.268 | 0.204 | 0.496 | 0.891 |

↑ the higher the better, ↓ the lower the better

Fig. 9. Benchmarking results for **CLINC150** when best checkpoint of the model is used during training and OOD label is not present in the training data.

| Feature Extractor | Detector | use_best_ckpt = True, is_ood_label_in_train = False | | | | | |
|---|---|---|---|---|---|---|---|
| | | F1↑ | MCC↑ | FPR@95↓ | FPR@90↓ | AUPR↑ | AUROC↑ |
| BERT | DOC | 0.497 | 0.502 | 0.875 | 0.750 | 0.691 | 0.793 |
| | Entropy | 0.703 | 0.658 | 0.708 | 0.492 | 0.808 | 0.875 |
| | LOF | 0.027 | 0.084 | 0.271 | 0.223 | 0.532 | 0.859 |
| | MSP | 0.393 | 0.420 | 0.898 | 0.795 | 0.617 | 0.750 |
| | TrustScores | 0.477 | 0.259 | 0.057 | 0.036 | 0.922 | 0.982 |
| MPNET | DOC | 0.621 | 0.601 | 0.747 | 0.494 | 0.831 | 0.890 |
| | Entropy | 0.700 | 0.659 | 0.489 | 0.387 | 0.820 | 0.894 |
| | LOF | 0.003 | 0.035 | 0.303 | 0.265 | 0.471 | 0.829 |
| | MSP | **0.706** | **0.677** | 0.612 | 0.224 | 0.885 | 0.928 |
| | TrustScores | 0.459 | 0.214 | **0.036** | **0.020** | **0.960** | **0.991** |
| USE | DOC | 0.399 | 0.428 | 0.877 | 0.753 | 0.678 | 0.789 |
| | Entropy | 0.551 | 0.506 | 0.729 | 0.481 | 0.687 | 0.815 |
| | LOF | 0.017 | 0.058 | 0.341 | 0.291 | 0.467 | 0.819 |
| | MSP | 0.395 | 0.415 | 0.903 | 0.805 | 0.596 | 0.736 |
| | TrustScores | 0.494 | 0.295 | 0.053 | 0.029 | 0.955 | 0.986 |

↑ the higher the better, ↓ the lower the better

Fig. 10. Benchmarking results for **ROSTD-Coarse** when best checkpoint of the model is used during training and OOD label is not present in the training data.

| Feature Extractor | Detector | use_best_ckpt = True, is_ood_label_in_train = False ||||||
|---|---|---|---|---|---|---|---|
| | | F1↑ | MCC↑ | FPR@95↓ | FPR@90↓ | AUPR↑ | AUROC↑ |
| BERT | DOC | 0.578 | 0.530 | 0.279 | 0.218 | 0.719 | 0.926 |
| | Entropy | 0.713 | 0.661 | 0.191 | 0.115 | 0.737 | 0.950 |
| | LOF | 0.514 | 0.420 | 0.536 | 0.413 | 0.467 | 0.843 |
| | MSP | 0.565 | 0.480 | 0.354 | 0.256 | 0.604 | 0.895 |
| | TrustScores | 0.268 | 0.046 | 0.160 | 0.133 | 0.645 | 0.935 |
| MPNET | DOC | 0.564 | 0.512 | 0.417 | 0.322 | 0.606 | 0.890 |
| | Entropy | 0.592 | 0.549 | 0.349 | 0.292 | 0.690 | 0.908 |
| | LOF | 0.000 | 0.000 | 0.572 | 0.418 | 0.318 | 0.801 |
| | MSP | 0.440 | 0.342 | 0.532 | 0.444 | 0.411 | 0.797 |
| | TrustScores | 0.268 | 0.046 | 0.182 | 0.120 | 0.768 | 0.957 |
| USE | DOC | 0.434 | 0.423 | 0.736 | 0.473 | 0.601 | 0.860 |
| | Entropy | 0.471 | 0.381 | 0.511 | 0.406 | 0.499 | 0.835 |
| | LOF | 0.561 | 0.490 | 0.320 | 0.297 | 0.592 | 0.890 |
| | MSP | 0.380 | 0.394 | 0.883 | 0.771 | 0.506 | 0.766 |
| | TrustScores | 0.268 | 0.046 | 0.287 | 0.211 | 0.668 | 0.928 |

↑ the higher the better, ↓ the lower the better

Fig. 11. Benchmarking results for **SNIPS** when best checkpoint of the model is used during training and OOD label is not present in the training data.

| Feature Extractor | Detector | use_best_ckpt = True, is_ood_label_in_train = False ||||||
|---|---|---|---|---|---|---|---|
| | | F1↑ | MCC↑ | FPR@95↓ | FPR@90↓ | AUPR↑ | AUROC↑ |
| BERT | DOC | 0.435 | 0.096 | 0.900 | 0.824 | 0.366 | 0.607 |
| | Entropy | 0.099 | 0.065 | 0.929 | 0.877 | 0.343 | 0.574 |
| | LOF | 0.446 | 0.135 | 0.896 | 0.770 | 0.313 | 0.594 |
| | MSP | 0.418 | 0.032 | 0.928 | 0.877 | 0.291 | 0.524 |
| | TrustScores | 0.427 | 0.059 | 0.898 | 0.821 | 0.285 | 0.543 |
| MPNET | DOC | 0.499 | 0.261 | 0.699 | 0.616 | 0.411 | 0.708 |
| | Entropy | 0.500 | 0.279 | 0.747 | 0.671 | 0.473 | 0.719 |
| | LOF | 0.462 | 0.278 | 0.799 | 0.743 | 0.476 | 0.705 |
| | MSP | 0.513 | 0.295 | 0.687 | 0.585 | 0.426 | 0.715 |
| | TrustScores | 0.515 | 0.294 | 0.731 | 0.609 | 0.343 | 0.670 |
| USE | DOC | 0.484 | 0.234 | 0.784 | 0.686 | 0.420 | 0.690 |
| | Entropy | 0.517 | 0.307 | 0.773 | 0.673 | 0.503 | 0.733 |
| | LOF | 0.281 | 0.162 | 0.872 | 0.739 | 0.416 | 0.669 |
| | MSP | 0.464 | 0.182 | 0.775 | 0.706 | 0.485 | 0.683 |
| | TrustScores | 0.497 | 0.265 | 0.717 | 0.626 | 0.320 | 0.640 |

↑ the higher the better, ↓ the lower the better

Fig. 12. Benchmarking results for **YELLOW** when best checkpoint of the model is used during training and OOD label is not present in the training data.

| Feature Extractor | Detector | use_best_ckpt = True, is_ood_label_in_train = True | | | | | |
|---|---|---|---|---|---|---|---|
| | | F1↑ | MCC↑ | FPR@95↓ | FPR@90↓ | AUPR↑ | AUROC↑ |
| BERT | DOC | 0.327 | 0.372 | 0.737 | 0.582 | 0.578 | 0.801 |
| | MSP | 0.428 | 0.457 | 0.501 | 0.316 | 0.718 | 0.887 |
| MPNET | DOC | 0.476 | 0.492 | 0.671 | 0.466 | 0.687 | 0.858 |
| | MSP | 0.703 | 0.687 | **0.204** | **0.111** | **0.877** | **0.960** |
| USE | DOC | 0.342 | 0.411 | 0.698 | 0.573 | 0.687 | 0.844 |
| | MSP | **0.714** | **0.698** | 0.291 | 0.152 | 0.876 | 0.951 |

↑ the higher the better, ↓ the lower the better

Fig. 13. Benchmarking results for **CLINC150** when best checkpoint of the model is used during training and OOD label is present in the training data.

| Feature Extractor | Detector | use_best_ckpt = True, is_ood_label_in_train = True | | | | | |
|---|---|---|---|---|---|---|---|
| | | F1↑ | MCC↑ | FPR@95↓ | FPR@90↓ | AUPR↑ | AUROC↑ |
| BERT | DOC | 0.994 | 0.992 | 0.000 | 0.000 | 1.000 | 1.000 |
| | MSP | 0.996 | 0.995 | 0.000 | 0.000 | 1.000 | 1.000 |
| MPNET | DOC | 0.998 | 0.997 | **0.000** | **0.000** | **1.000** | **1.000** |
| | MSP | **0.999** | **0.998** | 0.000 | 0.000 | 1.000 | 1.000 |
| USE | DOC | 0.996 | 0.994 | 0.000 | 0.000 | 0.999 | 1.000 |
| | MSP | 0.998 | 0.998 | 0.000 | **0.000** | 1.000 | 1.000 |

↑ the higher the better, ↓ the lower the better

Fig. 14. Benchmarking results for **ROSTD-Coarse** when best checkpoint of the model is used during training and OOD label is present in the training data.

| Feature Extractor | Detector | use_best_ckpt = True, is_ood_label_in_train = True | | | | | |
|---|---|---|---|---|---|---|---|
| | | F1↑ | MCC↑ | FPR@95↓ | FPR@90↓ | AUPR↑ | AUROC↑ |
| BERT | DOC | **0.897** | **0.878** | 0.039 | 0.017 | 0.970 | 0.994 |
| | MSP | 0.873 | 0.852 | 0.037 | 0.034 | 0.957 | 0.991 |
| MPNET | DOC | 0.882 | 0.861 | 0.037 | 0.020 | 0.968 | 0.994 |
| | MSP | 0.880 | 0.863 | **0.032** | **0.015** | **0.971** | **0.995** |
| USE | DOC | 0.858 | 0.833 | 0.047 | 0.030 | 0.952 | 0.991 |
| | MSP | 0.871 | 0.852 | 0.051 | 0.020 | 0.961 | 0.993 |

↑ the higher the better, ↓ the lower the better

Fig. 15. Benchmarking results for **SNIPS** when best checkpoint of the model is used during training and OOD label is present in the training data.

| Feature Extractor | Detector | use_best_ckpt = True, is_ood_label_in_train = True | | | | | |
|---|---|---|---|---|---|---|---|
| | | F1↑ | MCC↑ | FPR@95↓ | FPR@90↓ | AUPR↑ | AUROC↑ |
| BERT | DOC | 0.441 | 0.119 | 0.867 | 0.801 | 0.323 | 0.559 |

| Feature Extractor | Detector | use_best_ckpt = True, is_ood_label_in_train = True ||||||
| | | *F1*↑ | *MCC*↑ | *FPR@95*↓ | *FPR@90*↓ | *AUPR*↑ | *AUROC*↑ |
| | MSP | 0.428 | 0.098 | 0.872 | 0.784 | 0.326 | 0.565 |
| MPNET | DOC | 0.495 | 0.250 | 0.791 | 0.698 | 0.366 | 0.656 |
| | MSP | 0.490 | 0.240 | 0.705 | 0.642 | 0.393 | 0.680 |
| USE | DOC | 0.456 | 0.167 | 0.831 | 0.766 | 0.347 | 0.633 |
| | MSP | 0.484 | 0.243 | 0.734 | 0.663 | 0.362 | 0.663 |

↑ the higher the better, ↓ the lower the better

Fig. 16. Benchmarking results for **YELLOW** when best checkpoint of the model is used during training and OOD label is present in the training data.